\documentclass{article}
\usepackage{ijcai22}

\usepackage[T1]{fontenc}
\usepackage{aecompl}

\usepackage{times}
\usepackage{soul}
\usepackage{url}
\usepackage[colorlinks,
            linkcolor=blue,
            anchorcolor=blue,
            citecolor=blue,
            ]{hyperref}
\usepackage[utf8]{inputenc}
\usepackage[small]{caption}
\usepackage{graphicx}
\usepackage{amsmath}
\usepackage{amsthm}
\usepackage{booktabs}
\usepackage{algorithmicx}  
\usepackage{amsfonts}
\usepackage{xcolor}
\usepackage{amssymb}
\usepackage{arydshln}

\usepackage{url}

\newcommand{\tabincell}[2]{\begin{tabular}{@{}#1@{}}#2\end{tabular}}
\usepackage{multirow}

\usepackage[ruled,vlined]{algorithm2e}%[ruled,vlined]{
\usepackage{algpseudocode}

\newcommand{\tablesize}{\fontsize{7.5pt}{\baselineskip}\selectfont}
\usepackage[misc]{ifsym}
\newcommand\blfootnote[1]{%
\begingroup
\renewcommand\thefootnote{}\footnote{#1}%
\addtocounter{footnote}{-1}%
\endgroup
}

\title{BiFSMN: Binary Neural Network for Keyword Spotting}

\author{
Haotong Qin$^{1,2 *}$\and 
Xudong Ma$^{1 *}$\and 
Yifu Ding$^{1 *}$\and 
Xiaoyang Li$^2$\and \\
Yang Zhang$^2$\and
Yao Tian$^2$\and
Zejun Ma$^2$\and 
Jie Luo$^1$\And 
Xianglong Liu$^{1 \dag}$
\affiliations
$^1$Beihang University \\
$^2$Bytedance AI Lab
\emails
\{qinhaotong,yifuding,luojie,xlliu\}@buaa.com\quad
macaron\_lin@outlook.com
\{lixiaoyang.x,zhangyang.elfin,tianyao.11,mazejun\}@bytedance.com
}

\begin{document}

\maketitle

\begin{abstract}
The deep neural networks, such as the Deep-FSMN, have been widely studied for keyword spotting (KWS) applications. However, computational resources for these networks are significantly constrained since they usually run on-call on edge devices. 
In this paper, we present \textbf{BiFSMN}, an accurate and extreme-efficient binary neural network for KWS.
We first construct a \textit{High-frequency Enhancement Distillation} scheme for the binarization-aware training, which emphasizes the high-frequency information from the full-precision network's representation that is more crucial for the optimization of the binarized network.
Then, to allow the instant and adaptive accuracy-efficiency trade-offs at runtime, we also propose a \textit{Thinnable Binarization Architecture} to further liberate the acceleration potential of the binarized network from the topology perspective. 
Moreover, we implement a \textit{Fast Bitwise Computation Kernel} for BiFSMN on ARMv8 devices which fully utilizes registers and increases instruction throughput to push the limit of deployment efficiency.
Extensive experiments show that BiFSMN outperforms existing binarization methods by convincing margins on various datasets and is even comparable with the full-precision counterpart.
% (editor) (\textit{e.g.}, less than 3\% drop on Speech Commands V1-12). 
We highlight that benefiting from the thinnable architecture and the optimized 1-bit implementation, BiFSMN can achieve an impressive $\mathbf{22.3\times}$ speedup and $\mathbf{15.5\times}$ storage-saving on real-world edge hardware.
Our code is released at \url{https://github.com/htqin/BiFSMN}.
\end{abstract}

\section{Introduction}
With the advent of deep neural networks that process speech, great success has been achieved on speech tasks, such as the Deep Feedforward Sequential Memory Networks (D-FSMN)~\cite{zhang2018deep}.
\blfootnote{$^*$ equal contribution \quad $^\dag$ corresponding author}
And the deep neural networks for keyword spotting (KWS) are becoming more widely studied for real-world applications on the Internet of Things devices~\cite{chen2014small,zhang2015feedforward,ren2020fastspeech}, which enables users to activate devices by speaking keywords or specific phrases, while keeping the device sleeping when inactive. %but responsive at any time. 
However, such networks are usually deployed on edge devices with limited computation and power while running on-call to wait for any possible speech, which makes the resources for KWS always seriously constrained. 
To address the challenge, novel algorithms, such as cFSMN~\cite{chen2018compact}, DC-CNN~\cite{zhang2017hello}, and BC-ResNet~\cite{kim2021broadcasted}, have been proposed for efficient deep learning for KWS. 
Although these works have achieved remarkable speedup and memory footprint reduction, they still rely on expensive floating-point operations.
% (editor) , leaving room for further compression from model quantization perspective. 

With the most aggressive bit-width, model binarization~\cite{XNOR,qin2022bibert} emerges as one of the most promising quantization approaches to compress networks for better computational and storage-usage efficiency. Binary neural networks leverage (1) compact binarized parameters that take limited storage and (2) highly efficient bitwise operations far less costly than their floating-point counterparts. 
However, although the model binarization community has progressed, binarizing the network for KWS by existing methods is still far from ideal.
First, since the application of 1-bit parameters, the representation space of the binarized network is extremely limited and hard to optimize. The direct usage of existing binarization methods severely damages the accuracy of binarized models.
Second, existing architectures for KWS have fixed model scales and topologies, which cannot adaptively balance the resource budgets at runtime. %For the existing architecture after binarized, the room for optimization still exists from the orthogonal topology perspective, and the directly binarized networks are also hard to adaptively satisfies different resource constraints. 
Moreover, existing deployment frameworks are far from reaching the theoretical upper limit of acceleration for the binarized network when implemented on real-world hardware. 

\begin{figure*}[!ht]
\centering
% \vspace{-0.5in}
\includegraphics[width=16cm]{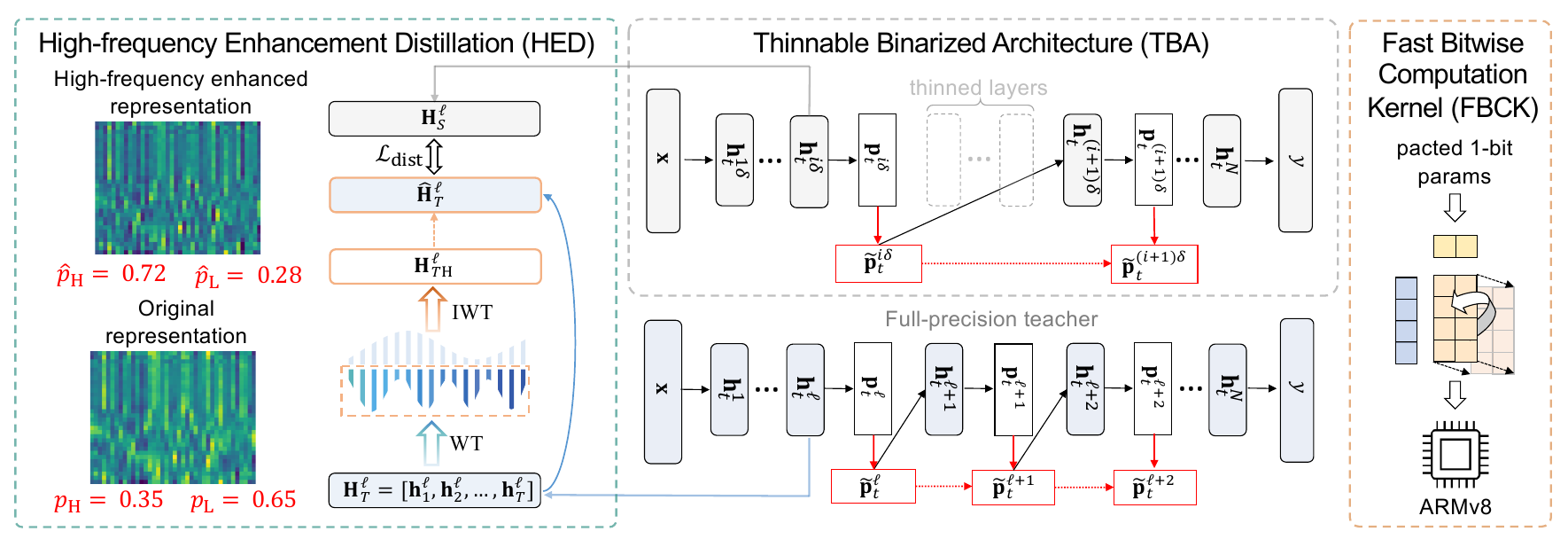}
% \vspace{-0.17in}
\caption{Overview of our BiFSMN, which applies High-frequency Enhancement Distillation (HED) to emphasize significant features and Thinnable Binarized Architecture (TBA) to balance accuracy-efficiency trade-offs, and is implemented by Fast Bitwise Computation Kernel (FBCK) for efficient deployment on real-world devices.} 
\label{fig:overview}
\end{figure*}

This paper presents a \textbf{Bi}narized \textbf{F}eedforward \textbf{S}equential \textbf{M}emory \textbf{N}etwork (\textbf{BiFSMN}), which emerges as a successful practice of binary network for KWS application (see the overview in Figure~\ref{fig:overview}), with accurate prediction, lightweight computation, and efficient deployment.
BiFSMN is built based on the binarization of D-FSMN, which is a pure feedforward structure.
First, we construct a \textit{High-frequency Enhancement Distillation} (HED) scheme for binarization-aware training, which emphasizes high-frequency information of the full-precision teacher's representation via wavelet transform that is more crucial for the optimization of the binarized network. % and makes the significant information easier to be learned by the binarized student.
Second, to enable adaptive accuracy-efficiency trade-offs at runtime, we propose a \textit{Thinnable Binarization Architecture} (TBA) to build various depths (such as $[1.0, 0.5, 0.25]\times$), which further liberates the acceleration potential of the binarized network from the topology perspective to satisfy the different resource constraints. % which consists of thinnable binarized blocks
% The proposed BiFSMN keeps accuracy even with smaller model sizes. %The proposed BiFSMN is trained by knowledge distillation via the uniform layer-mapping strategy to keep accuracy even with smaller model sizes. 
Moreover, we also provide an efficient \textit{Fast Bitwise Computation Kernel} (FBCK) for BiFSMN on ARMv8 devices. On real-world hardware, the implementation of the BiFSMN using FBCK enjoys impressively faster inference than that using the existing binarization frameworks. 

Our BiFSMN is the first binary neural network specialized for KWS.
Extensive experiments on Google Speech Commands V1 and V2 datasets (12, 20, and 35 classification tasks) show that our BiFSMN completely outperforms existing binarization methods and is even almost accurate on par with the full-precision counterpart, \textit{e.g.}, BiFSMN just drops within 3\% on Speech Commands V1-12. %And BiFSMN also outperforms existing binarization methods on the same architecture by convincing margins.
Besides, we highlight that the efficient implementation makes our BiFSMN easy deployment and fast inference in real-world devices: in actual evaluation on edge ARM devices, BiFSMN can achieve up to $22.3\times$ speedup and $15.5\times$ storage-saving compared with the full-precision D-FSMN.

\section{Related Work}

\subsection{Network Binarization}
\label{sec:NetworkBinarization}
Recently, various binarization methods for neural networks have emerged to compress and accelerate networks. 
The existing binarization methods are designed to obtain accurate binarized networks by minimizing the quantization error~\cite{XNOR}, improving loss function~\cite{Regularize-act-distribution}, \textit{etc}. 
And from the architectures prospective, despite the most popular CNNs~\cite{BiReal}, transformer-based and MLP-based networks are also studied for binarization~\cite{qin2020bipointnet}.
% (editor) Unfortunately, we show in Section~\ref{sec:Experiments} that existing binarization methods are not readily transferable to build accurate binary networks for KWS.
The practical use of binarization on real devices relies on deployment support. There are binarization frameworks with different target platforms (\textit{e.g.}, CPUs, GPUs, and FPGAs) and applications, such as daBNN~\cite{zhang2019dabnn} and Bort~\cite{shang2021hw}. 
% (editor) However, these binarized networks implemented by these frameworks suffer much lower efficiency than the theoretical performance. 

\subsection{Deep Learning for Keyword Spotting}
Due to their learning potential and superior performance, deep neural networks for KWS have become widely studied. 
One classic model is the recurrent neural network (RNN), which enables to capture of the context in a sequence of data~\cite{9414339}.
CNN-based (BC-ResNet~\cite{kim2021broadcasted}) and transform-based (Audiomer) models for KWS are also proposed for better performance and cheaper energy cost. 
Feedforward Sequential Memory Networks (FSMN)~\cite{zhang2015feedforward} mitigates the vanishing gradient problem of RNNs, and is also efficient in computation in convergence. 
Improvements on the original design of FSMN are varied, such as compact FSMN (cFSMN)~\cite{chen2018compact}, Deep-FSMN (D-FSMN), and pyramidal FSMN (pFSMN)~\cite{yang2018novel}.
% (editor) However, they still use expensive floating-point parameters and operations, which can be improved by binarization.

\section{BiFSMN}

In this section, we present the BiFSMN for KWS application.
We first build a basic binarization framework and then introduce our techniques, including \textit{High-frequency Enhancement Distillation} (HED), \textit{Thinnable Binarization Architecture} (TBA), and \textit{Fast Bitwise Computation Kernel} (FBCK). 

%\subsection{Binarized D-FSMN Architecture}
\subsection{Basic Binarization Framework}
\label{subsec:BinarizedD-FSMNArchitecture}
%\subsubsection{Binarization}

Here we give an introduction to the process of obtaining the basic binarization framework.
As one of the most classic and widely used models for speech tasks, D-FSMN~\cite{zhang2018deep} is considered as a binarization-friendly architecture for the following reasons:
(1) The D-FSMN is a pure feedforward structure with the backbone built up by stacking FIR-like memory blocks.
(2) D-FSMN introduces skip connections between memory blocks in adjacent layers to allow the information to flow directly to the next layer, which is also proved important for accurate binarized networks~\cite{BiReal}.
Therefore, we consider the construction of the binarized D-FSMN as the basic binarization framework.

%\subsubsection{Binarization Function}
We first introduce the basic formulations of binarization.
In the binarized network, both weights and activations are compressed to 1-bit using the $\operatorname{sign}$ function in the forward propagation, and the $\operatorname{STE}$~\cite{courbariaux2015binaryconnect} is applied to clip the gradient in the backward propagation:
\begin{equation}
\small
\label{eq:quantized}
\operatorname{sign}(x)=
\begin{cases}
1& \text{if } x \ge 0\\
-1& \text{otherwise }
\end{cases},
\frac{\partial C}{\partial x}=
\begin{cases}
\frac{\partial C}{\partial\operatorname{sign}(x)}& \text{if } |x| \leq 1\\
0& \text{otherwise }
\end{cases},
\end{equation}
where $C$ is the cost function for the minibatch, $x$ denotes the element in floating-point parameters.
The scaling factor $\alpha$ is also introduced to retain the magnitude of real-value weights:
\begin{equation}
\label{eq:bwn}
\alpha_{\mathbf{w}} = \frac{1}{n}\mathbf{w}^\top \mathtt{sign} (\mathbf{w}) = \frac{1}{n} \left\| \mathbf{w} \right\|_{l1},
\end{equation}
we make $\mathbf{w} \approx \alpha_{\mathbf{w}} \mathbf{B_w}$ to reduce the quantization error, where $\mathbf{B_w}$ is the 1-bit weight binarized by $\operatorname{sign}$ function.

% With the challenge of limited representation capability and difficulty in training 1-bit parameters, we propose a novel binarized FSMN, dubbed BiFSMN. 

%\subsubsection{Binarized Architecture}
%Different from the most common CNNs, networks like FSMN have a special characteristic that it is built up by stacking several layer blocks. Each block can be described as a same memorable structure which merges the standard feedforward sequantial neuron and its memory unit. When constructing a binarized FSMN, we replace both the floating-point weight matrix and the hidden states by binarized parameters:

Then, we introduce the binarized network architecture.
When constructing a binarized D-FSMN, we binarize both the floating-point weight and activation for linear and convolutional layers.
Given the input $\ell$-th hidden states $\mathbf{H}^\ell=[\mathbf{h}_1^\ell, \mathbf{h}_2^\ell, ..., \mathbf{h}_T^\ell]$, and each $\mathbf{h}_t^\ell; t\in[1, T]$ denotes a fixed-size representation of the long surrounding context at time instance $t$.
The formulation of the $\ell$-th binarized memory block takes the following form:
\begin{equation}
\begin{aligned}
\tilde{\mathbf{p}}_{t}^{\ell}=&\sum_{i=0}^{N_{1}^{\ell}} \alpha_{\mathbf{a}_i}^\ell\left(\mathbf{B}_{\mathbf{a}_{i}}^{\ell} \otimes \mathbf{B_p}_{t-is_{1}}^{\ell}\right)\\
+&\sum_{j=1}^{N_{2}^{\ell}} \alpha_{\mathbf{c}_j}^\ell\left(\mathbf{B}_{\mathbf{c}_{j}}^{\ell} \otimes \mathbf{B_p}_{t+js_{2}}^{\ell}\right)+\mathcal{H}\left(\tilde{\mathbf{p}}_{t}^{\ell-1}\right)+\mathbf{p}_{t}^{\ell}.
\end{aligned}
\end{equation}
Here, $\mathbf{p}_{t}^{\ell}=\alpha_\mathbf{V}\left(\mathbf{B_V}^{\ell}\otimes\mathbf{B_h}_{t}^{\ell}\right)+\mathbf{b}^{\ell}$ denotes the linear output of the binarized linear projection layer. $\otimes$ denotes the inner product with bitwise operations XNOR and Bitcount.
$\tilde{\mathbf{p}}_{t}^{\ell}$ denotes the output of the memory block.
$\mathcal{H}(\cdot)$ denotes the skip connection (identity mapping) within the memory block. $N_{1}^{\ell}$ and $N_{2}^{\ell}$ denotes the look-back and lookahead orders and $s_1$ and $s_2$ are the stride for look-back and lookahead filters respectively. 
The input of the units for the next hidden layer $\mathbf{H}^{\ell+1}=[\mathbf{h}_1^{\ell+1}, \mathbf{h}_2^{\ell+1}, \cdots, \mathbf{h}_T^{\ell+1}]$ is calculated as follow:
\begin{equation}
\label{eq:4}
\mathbf{h}_{t}^{\ell+1}=f\left(\alpha_\mathbf{U}^{\ell}\left(\mathbf{B}_\mathbf{U}^{\ell}\otimes \tilde{\mathbf{B}}_{\mathbf{p}_t}^{\ell}\right)+\mathbf{b}^{\ell+1}\right),
\end{equation}
where $f(\cdot)=\operatorname{BN}\cdot\operatorname{Nonlinear}(\cdot)$ denotes the composition of batch normalization and nonlinear functions (PReLU in the binarized network~\cite{RealtoBin}).

\subsection{High-frequency Enhancement Distillation for Binarization-aware Training}

\begin{figure}[t]
\centering
% \vspace{-0.5in}
\includegraphics[width=7cm]{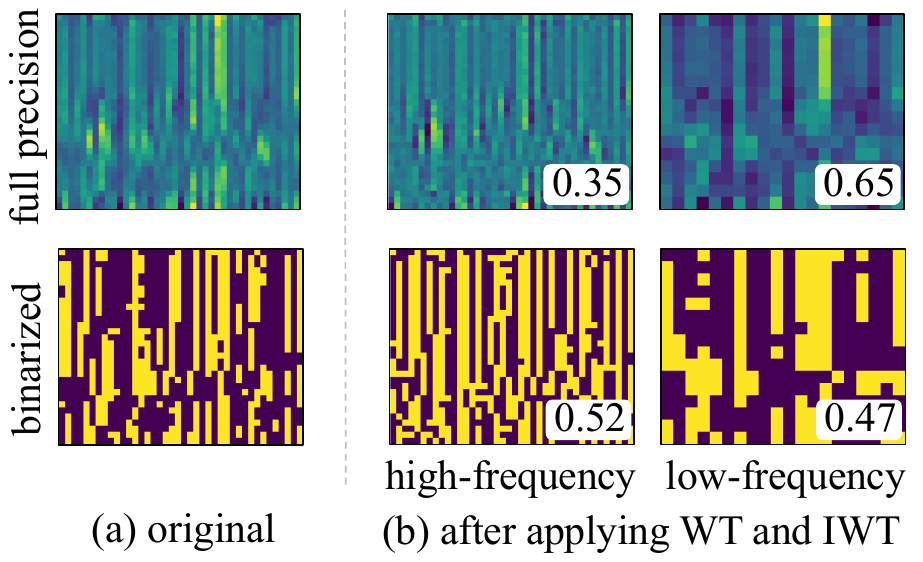}
% \vspace{-0.15in}
\caption{%Full-precision and binarized features before and after wavelet transform. 
Compared with the original full-precision representation, the relative wavelet energy of the high-frequency component of the binarized representation is larger while that of the low-frequency component is much smaller.}
\label{fig:wavelet}
\end{figure}

When the neural network is binarized, its representation capability is extremely limited, and the accuracy is significantly decreased compared with the full-precision counterparts. As shown in Figure~\ref{fig:wavelet}, binarization makes the intermediate representation visually monotonous since it restricts the pixels to binary, while the original feature information is richer and contains much detailed information.
% (editor) Therefore, some existing binarization works optimize the binarized representation globally in the training process, such as applying entropy-maximized or learnable mean-shift to pre-binarized weight or activation~\cite{qin2019forward,liu2020reactnet}.
However, since it is highly discrete in binarized representations, values on the edges falling to 1 or -1 outline the object, conveying more important information than plain blocks. While it is well-known that edges often present as local maxima of the gradient, which is hard to be improved directly by global optimization.

Fortunately, we find that the essence of the information inclination to edge is that the essential binarized representation tends to concentrate on high-frequency components.
We use 2D Haar Wavelet Transform (WT)~\cite{meyer1992wavelets}, which is most frequently used as a separable transform that isolates horizontal and vertical edges, to decompose the representation into low and high-frequency components. 
The hidden state $\mathbf{H}$ inputted to a specific layer can be represented as a weighted sum of the wavelet function family:
\begin{equation}
\label{eq:haar}
%\textrm{DWT}: F(t) = \sum^{-1}_{j=-N} \sum_k C_j(k) \phi_{j,k}(t),
f_\textrm{WT}(\mathbf{H}) = \sum^{-1}_{j=-N} \sum_k \mathbf{C}_j(k) \phi_{j,k},
\end{equation}
where $\phi$ is the mother wavelet function with a specific time parameter, $j=-1, \cdots, -N$ is the resolution level, and $k$ determines the translation of the waveform.
% 具体 haar wavelet 公式？
%Therefore, we formulate $C_j$ as the output of forward wavelet transform:
%\begin{equation}
%C_j = f_\textrm{DWT}(\mathbf{F}, t; \phi_{j}(t))
%\end{equation}
%We apply a high-pass filter and inverse wavelet transform to get new features where only high-pass coefficients are reserved. The process can be formulated as follow:
%\begin{equation}
%\mathbf{F}^{h} = f_\textrm{DWT}^{-1}(\sum_j C_j^h; \phi_{j}(t))
%\end{equation}

%Then, to measure the information amount conveyed by the single component of representation, we introduce the wavelet entropy~\cite{rosso2001wavelet}, which quantifies the energy distribution in wavelet sub-frequency bands, and the value quantitatively reflects the series' complexity. 
To measure the information amount conveyed by the single component of representation, the relative wavelet energy is used to define the amount of information~\cite{rosso2001wavelet}.
The wavelet energy $E_j$ at $j$-th level is first calculated as:
\begin{equation}
\label{eq:wavelet-energy}
E_j = \sum_k | \mathbf{C}_j(k) |^2.
\end{equation}
%When the total resolution levels is set as $N=2$ intuitively, we have $\mathbf{C}_j\in[\mathbf{C}_\textrm{L}, \mathbf{C}_\textrm{H}]$, where $\mathbf{C}_\textrm{L}$ and $\mathbf{C}_\textrm{H}$ denote the low and high-frequency coefficients, respectively. 
When we obtain low and high-frequency coefficients ($\mathbf{C}_j\in\{\mathbf{C}_\textrm{L}, \mathbf{C}_\textrm{H}\}$, $N=2$) by once decomposition, their relative wavelet energy $p_\textrm{H}$ and $p_\textrm{L}$ can be expressed as:
\begin{equation}
p_\textrm{H}=\frac{E_\textrm{H}}{E_\textrm{H}+E_\textrm{L}},\qquad p_\textrm{L}=\frac{E_\textrm{L}}{E_\textrm{H}+E_\textrm{L}}.
\end{equation}
%And the total energy can be obtained by summing up the energy during each sampled time series $k$ at each decomposition level $j$:
%\begin{equation}
%\label{eq:total-energy}
%E_\textrm{tot} = E_\textrm{L} + E_\textrm{H} = \sum_k | C_\textrm{L}(k) |^2 + | C_\textrm{H}(k) |^2.
%\end{equation}
%And the total wavelet energy is defined as:
%\begin{equation}
%\label{eq:total-wavelet-entropy}
%S_\textrm{WT} = S_\textrm{H} + S_\textrm{L} = \frac{E_\textrm{H}}{E_\textrm{tot}}\ \textrm{ln}\left(\frac{E_\textrm{H}}{E_\textrm{tot}}\right)+\frac{E_\textrm{L}}{E_\textrm{tot}}\ \textrm{ln}\left(\frac{E_\textrm{L}}{E_\textrm{tot}}\right),
%\end{equation}
%where the total energy $E_\textrm{tot} = E_\textrm{L} + E_\textrm{H}$. 
The larger relative wavelet energy demonstrates the information is more gathered in this component.
%We present the wavelet entropy and visualization of low and high-frequency components of the intermediate feature in Figure~\ref{fig:wavelet}.
%As Figure~\ref{fig:wavelet} shows, compared with the full-precision features, the wavelet entropy of the high-frequency components of the binarized features increases significantly, while that of the low-frequency components decreases. The phenomenon implies that binarized representation inclines higher-frequency components. 
As Figure~\ref{fig:wavelet} shows, compared with the full-precision representation, the relative wavelet energy of the high-frequency components of the binarized ones increases significantly, which implies that binarized representation inclines higher-frequency components.

% Based on the above analysis, we propose a High-frequency Enhancement Distillation for binarization-aware training.
% The scheme utilizes a pre-trained full-precision D-FSMN as the teacher, and enhances the high-frequency components of its hidden states during distillation.
% Specifically, we apply wavelet transform on the original representations, remove the low-frequency component, and then apply Inverse Wavelet Transform (IWT) function on the high-frequency components to retrieve selected representation.
We propose a High-frequency Enhancement Distillation according to the aforementioned analysis. The scheme exploits a full-precision pre-trained D-FSMN model as the teacher to assist the binarization-aware training by enhances the high-frequency components of the full-precision representation in hidden layers. 
We first apply the wavelet transform on the original hidden states 
The process can be formulated as follow:
\begin{equation}
\mathbf{H}_{T\textrm{H}} = f_\textrm{IWT}\left(\sum_k \mathbf{C}_{T\textrm{H}}(k) \phi_{T\textrm{H},k}\right).
\end{equation}
%We enhance the high-frequency information by adding up the original and emphasized high-frequency features as the distilled knowledge to provide better guidance in training binarized student network: 
And then the emphasized high-frequency representations are added to the original ones:
\begin{equation}
\hat{\mathbf{H}}_{T} = \frac{\mathbf{H}_{T\textrm{H}}}{\sigma(\mathbf{H}_{T\textrm{H}})} + \frac{\mathbf{H}_{T}}{\sigma(\mathbf{H}_{T})},
\end{equation}
where $\sigma(\cdot)$ is the standard deviation.
Then, inspired by \cite{RealtoBin}, we minimize the attention distillation loss between $\hat{\mathbf{H}}_{T}$ from the teacher and $\hat{\mathbf{H}}_{S}$ directly from the hidden states of the student, which is expressed as:
\begin{equation}
\label{eq:loss-r2b}
\mathcal{L}_\textrm{dist} = \sum^{N}_{\ell=1} \left\| \frac{\mathbf{H}^{\ell\ 2}_S}{\|\mathbf{H}^{\ell\ 2}_S\|} - \frac{\hat{\mathbf{H}}^{\ell\ 2}_T}{\| \hat{\mathbf{H}}^{\ell\ 2}_T\|} \right\|,
\end{equation}
where $\ell$ denotes the $\ell$-th block and $\|\cdot\|$ is the L2-norm.

The HED scheme above makes it easier for the binarized student network to exploit the essential information from emphasized full-precision representations and improve accuracy. 

\begin{figure}
\centering
% \vspace{-0.4in}
\includegraphics[width=6.5cm]{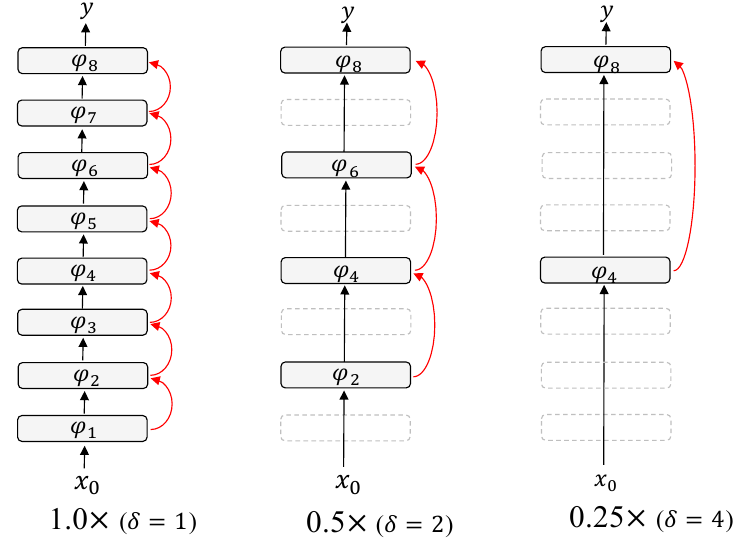}
% \vspace{-0.15in}
\caption{An instance of Thinnable Binarization Architecture ($N=8$, $\delta=1, 2, 4$) at runtime, where red arrows denote skip connections.}
\label{fig:thinnable}
\end{figure}

\subsection{Thinnable Binarization Architecture for Runtime Accuracy-efficiency Trade-off}

%As discussed earlier, binarization is an efficient approach to compress and speed up the neural network with extremely lightweight 1-bit parameters and efficient bitwise operations, enabling fast inferences on resource-limited devices. As one of the advantages, binarization will not affect the model architecture, which is critical for carefully engineered structures.
As discussed earlier, binarization is an efficient compression approach with extremely lightweight 1-bit parameters and efficient bitwise operations, enabling fast inferences on resource-limited devices. As one of the advantages, binarization will not affect the model architecture, which is critical for well-designed structures.
However, the energy budget varies between devices and even during wake-up and power-saving modes on a single device. 
Therefore, a lightweight yet adaptive binarized architecture, which can switch among different widths at runtime, permits instant and adaptive accuracy-efficiency trade-offs for KWS.

% As we discussed earlier, binarization aims to provide the neural network with extremely lightweight 1-bit parameters and efficient bitwise operations, which enables the binarized neural network to perform fast inferences on resource-limited devices. 
% And the binarized D-FSMN, chosen as the basic binarization architecture, mainly consists of several binarization-friendly blocks that include linear layers, short connections, \textit{etc}.
% The binarization-friendly basic architecture and High-frequency Enhancement Distillation make the accuracy of the binarized network even close to that of the full-precision counterpart.

% However, the binarized parameters may still be redundant, since the binarization affects the bit-width of parameters rather than the number. 
% Moreover, under the limit of ultra-low power consumption, the computational and energy budget varies for KWS applications even on the same device (for example, a device may be in low-power or power-saving mode).
% Thus, a lightweight yet adaptive binarized architecture for KWS is desired that can switch among different widths at runtime, permitting instant and adaptive accuracy-efficiency trade-offs.

We present a Thinnable Binarization Architecture (TBA) for KWS, which can select a thinner model with fewer layers at runtime, directly reducing the computational consumption.  
%The proposed thinnable binarized architecture consists of $N$ memory blocks ($N \in \{2^n, n\in\mathbf{Z}^+\}$) and requires an adaptive thinnable coefficient $\delta\in\mathbf{Z}\cap [1, \log_2 N]$.
When we focus on the computational expensive backbone, the basic binarization architecture $\textrm{M}'$ containing $N$ blocks ($N \in \{2^n, n\in\mathbf{Z}^+\}$) can be expressed as:
\begin{equation}
\textrm{M}'(\mathbf x) = \varphi^{N}\cdot\varphi^{N-1}\cdot...\cdot\varphi^1(\mathbf x),
\end{equation}
where $\textrm{M}'$ and $\varphi^\ell$ are the binarized network and $\ell$-th binarized D-FSMN block, respectively, and $\mathbf x$ is the input of network.
The TBA derived from $\textrm{M}$ can be defined as:
\begin{equation}
\textrm{M}(\mathbf x; \delta) = \Phi^{N}\cdot\Phi^{N-1}\cdot...\cdot\Phi^1(\mathbf x),
\end{equation}
where $\delta$ is the interval of selected layers, which is confined to be divisible into $N$.
And each thinnable block $\Phi^{\ell}$ can be defined as
\begin{equation}
\Phi^{\ell}(\mathbf x)=
\begin{cases}
\varphi^{\ell}(\mathbf x), & \ell\in \{i\delta, i \in[1, N/\delta]\},\\
\mathbf x, & \textrm{otherwise}.
\end{cases}
\end{equation}
And the batch normalization in the function $f(\cdot)$ in Eq.~(\ref{eq:4}) in $\ell$-th binarized D-FSMN block $\varphi^{\ell}$ are preassigned  according to different variants in the thinnable network. 
The thinnable network architecture will skip intermediate blocks every $\delta$ layers by replacing them with identity functions. 
% Figure~\ref{fig:overview} shows our thinnable binarization architecture and we takes the number of total layers $N=8$ in our experiments.
Figure~\ref{fig:overview} shows the formalization of our thinnable binarization architecture, and we also provide an instance for $N=8$, $\delta=1, 2, 4$ in Figure~\ref{fig:thinnable}, which is also the default setting in our experiments. 
%The thinnable binarized model permits instant and adaptive accuracy-efficiency trade-off at runtime.

%We adopt the uniform layer mapping strategy in distillation during training. 

% However, an apparent problem of shallow network is the learning ability.

\begin{algorithm}[t]
    \small
    \caption{The training process of our BiFSMN.}
    \label{alg:1}
    \KwIn{Fixed pre-trained full-precision teacher $\textrm{M}_\textrm{FP32}$ and thinnable binarized model $\textrm{M}$ (BiFSMN) with $N$ basic binarized blocks, training iterations $T$.}
    \KwOut{Well-trained thinnable binarized model $\textrm{M}$}
    %Initialize $\mathbf{x}^{0}$ from Gaussian distribution $\mathcal{N}(0,1)$\;
    %Forward propagate $\textrm{M}(\mathbf{x}^{0})$ and gather activations\;
    %Compute $\delta_i$ and $\gamma_i$ based on Eq.~(\ref{eq:mn})\;
    \For{all $t=1,2,\dots, T$}
    {
    Forward propagate $\textrm{M}_\textrm{FP}(\mathbf{x})$ and obtain the information-enhanced intermediate features $\hat{\mathbb{H}}_{T}=\{\hat{\mathbf{H}}_T^1, \hat{\mathbf{H}}_T^2, \dots, \hat{\mathbf{H}}_T^N\}$\;
    \For{all $\delta=1, 2, \dots, \log_2 N$}
    {
    Forward propagate $\textrm{M}(\mathbf{x}; \delta)$ and obtain the intermediate features $\mathbb{H}_{S}^\delta=\{\mathbf{H}_S^{\delta}, \mathbf{H}_S^{2\delta}, \dots, \mathbf{H}_S^{N}\}$\;
    Compute the distillation loss $\mathcal{L}_{\textrm{dist}}^\delta$ by Eq.~(\ref{eq:distill_loss_delta})\;
    Compute the cross-entropy loss $\mathcal{L}_{\textrm{CE}}^\delta$\;
    }
    Descend $\mathcal{L}_\textrm{tot}$ as Eq.~(\ref{eq:total_loss}) and update $\textrm{M}$\;
    }
    Get the well-trained BiFSMN model $\textrm{M}$\;
    Evaluate the BiFSMN on test dataset and get the accuracy.
\end{algorithm}

%\subsection{Training and Deployment}

% In this paper, High-frequency Enhancement Distillation is applied for the binarization-aware training to optimize thinnable binarization architecture, aiming to train a accurate and thinnable binarized network.

% For the distillation, we adopt the uniform layer mapping strategy, which is formulated as:
To optimize the binarization-aware training for the proposed TBA, we adopt the uniform layer mapping strategy to better align and learn representation in the HED: 
\begin{equation}
\label{eq:distill_loss_delta}
\mathcal{L}_{\textrm{dist}}^\delta = \sum^{N/\delta}_{i=1} \left\| \frac{\mathbf{H}^{i\delta\ 2}_S}{\| \mathbf{H}^{i\delta\ 2}_S\|} - \frac{\hat{\mathbf{H}}^{i\delta\ 2}_T}{\| \hat{\mathbf{H}}^{i\delta\ 2}_T\|} \right\|.
\end{equation}
% During backward propagation, the gradients from different switches are accumulated to jointly update the weight:
The gradients from different switches are accumulated during backward propagation to update the weight jointly. According to the compression ratio in thinnable architecture, the weighted loss can be calculated as: 
\begin{equation}
\label{eq:total_loss}
\mathcal{L}_{\textrm{tot}} = \sum_\delta \frac{1}{2^{\delta-1}} \left(\mathcal{L}_{\textrm{CE}}^\delta + \gamma \mathcal{L}_{\textrm{dist}}^\delta \right),
\end{equation}
where $\mathcal{L}_{\textrm{CE}}^\delta$ denotes the cross-entropy loss of $\mathrm{M}(\cdot; \delta)$ and $\gamma$ is a hyperparameter to control distillation impact, set to 0.01 as default. 
The detailed training procedures for the BiFSMN are listed in Algorithm~\ref{alg:1}.

% For deploying on devices with limited computing resources, we also implement the optimized 1-bit computation kernels to accelerate the inference on ARMv8-A architecture. 
% In the binarized convolution networks, the bitwise XNOR and bitcount operations are the core operations of Binarized General Matrix Multiply (BGEMM), which can be efficiently completed by the EOR, CNT, and ADD instructions on ARMv8. 
% In order to improve the local utilization of data and save memory handling, the block size of the core kernel of BGEMM is determined to be 4 (M) x 16 (N) according to the number of registers under the armv8 architecture; and after 16 times expands the accumulation of uint8 for K, we accumulate the data to the uint16-length register using a long instruction, which increases the computation width of the accumulation process; we also execute the tiling operation to the input according to the size of the cache. 

\subsection{Fast Bitwise Computation Kernel for Efficient Hardware Deployment}

Benefiting from binarized weights and activations compressed to $\frac{1}{32}$ of the original bit-width, a single binarized layer has an extreme-high $64\times$ theoretical reduction of FLOPs~\cite{BiReal}. However, when we implement and deploy the entire binary neural network on real-world hardware using existing binarization deployment frameworks, such as daBNN~\cite{zhang2019dabnn} and Bolt~\cite{bolt}, its overall inference efficiency is often significantly lower than the theoretical upper limit. One of the key bottlenecks of acceleration is the Binarized General Matrix Multiply (BGEMM) performed with the bitwise XNOR and Bitcount.
%(editor) In existing deployment frameworks like daBNN, the corresponding instructions on ARMv8 are \texttt{EOR}, \texttt{CNT}, \textit{etc}.
%(editor) However, owing to the limit of register and instruction throughput, the efficiency of BGEMM hardware implementation is far less than that of its theoretical design.
Therefore, for efficient deployment on edge devices with limited computational resources, we further optimize the 1-bit computation with new instruction and register allocation strategy to accelerate the inference on ARMv8-A architecture widely used on edge devices. We dub it Fast Bitwise Computation Kernel (FBCK). 

\begin{figure}
\centering
% \vspace{-0.4in}
\includegraphics[width=6cm]{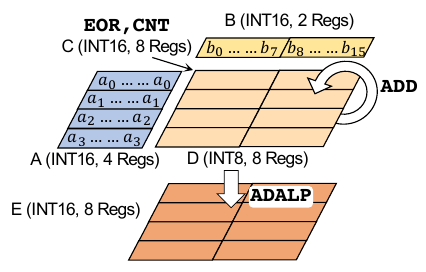}
% \vspace{-0.1in}
\caption{Fast Bitwise Computation Kernel for BiFSMN, which improves the utilization of registers to expand instruction throughput.}
\label{fig:deploy}
\end{figure}

According to the number of registers on ARMv8 architecture, we first reallocate the registers in the kernel as five partitions in order to improve the register utilization and reduce memory footprint: partition A has four registers (except register v0) for one input (weight/activation), B has two for the other input, C has eight for intermediate results of \texttt{EOR} and \texttt{CNT}, D has eight for the output in one loop, and E has eight for the final results. 
Each input is packed as INT16. Each register in A stores one input while repeated 8 times, while each in B stores 8 different inputs. We first apply \texttt{EOR} and \texttt{CNT} for A with one register of B to get 32 INT8 results in intermediate partition C, and then perform \texttt{ADD} to accumulate the INT8 to D, and do the same for the other register of B. 
After sixteen times loop, we finally accumulate the INT8 data stored in D to an INT16 register (in E) using long instruction \texttt{ADALP}, which extends INT8 data to double width. FBCK makes full use of registers almost without idle bits during the computation. Refer to Figure~\ref{fig:deploy} as illustration.

%We evaluate the performance of different ARMv8 devices, comp=ared with the open-source daBNN and Bolt binarization frameworks, our implementation achieves $3.7\times$ and $1.5\timesX$ speedup, respectively.

\section{Experiments}
\label{sec:Experiments}

%In this section, we showcase experiments to validate the effectiveness of the proposed BiFSMN. We first compare our HED method with representative binarization methods on several datasets, including the KWS task on the Google Speech Commands V1 and V2 datasets~\cite{}. We also demonstrate the versatility of the thinnable network to best adapt different resource constraints. Besides, we implement our high performance computation kernel on real devices to forcefully prove the potential in real application. 

In this section, we conduct experiments on the Google Speech Commands V1 and V2 datasets~\cite{warden2018speech} to verify the effectiveness of BiFSMN and compare it with state-of-the-art (SOTA) binarization methods and various architectures. 

% (editor) The BiFSMN for experiments has 8 memory blocks with 128 backbone memory size and 256 hidden size. Among the experiments, we apply SGD optimizer with 1e-4 weight decay and 5e-3 learning rate (decay to 0 using cosine annealing scheduler). For other networks and binarization methods, we mostly follow the hyper-parameter settings of their original papers. We train all binarized models 300 epochs.

\begin{table}[h]
\tablesize
\centering
\renewcommand\arraystretch{0.4}
% \vspace{-0.1in}
% \vspace{-0.15in}
% \setlength{\tabcolsep}{1.7mm}
{\begin{tabular}{llcrrr}
\toprule
{\bf Arch.} & {\bf Quant} & {\tabincell{c}{\textbf{\#Bits}}} & {\tabincell{r}{\textbf{FLOPs}}}  & {\tabincell{r}{\textbf{V1}}} & {\tabincell{r}{\textbf{V2}}}\\
\midrule
{\tabincell{l}{D-FSMN}} & Full Prec.    & 32/32 & 710.15 & 97.51 & 96.01 \\
\midrule
\multirow{3}{*}{\tabincell{l}{D-FSMN}} & Vanilla & 1/1 & 40.46 & 87.71 & 89.53 \\
& Distill & 1/1 & 40.46 & 90.02 & 90.95 \\
& HED & 1/1 & 40.46 & 93.47 & 93.54\\ 
\midrule
\multirow{12}{*}{\tabincell{l}{BiFSMN\\(TBA)\\$_{[1, 0.5, 0.25]\times}$}} & \multirow{3}{*}{Vanilla} & \multirow{3}{*}{1/1} & 40.46 & 87.72 & 89.96 \\
 & & & 29.90 & 86.95 & 88.85 \\
 & & & 24.62 & 84.19 & 87.09 \\
\cdashline{2-6}[1pt/1pt]
 & \multirow{3}{*}{Distill} & \multirow{3}{*}{1/1} & 40.46 & {92.31} & {93.00}  \\
 & & & 29.90 & 91.89 & 92.84 \\
 & & & 24.62 & 91.83 &  92.70 \\
\cdashline{2-6}[1pt/1pt]
 & \multirow{3}{*}{HED} & \multirow{3}{*}{1/1} & 40.46 & \textbf{95.03} & \textbf{94.86}  \\
 & & & 29.90 & 94.87 & 94.74 \\
 & & & \textbf{24.62} & 94.48 & 94.63 \\
\bottomrule
\end{tabular}}
\caption{Ablation study of BiFSMN on Speech Command 12 tasks. FLOPs denotes million FLOPs, same below. }
\label{tab:ablation}
\end{table}

\subsection{Ablation Study}

We perform ablation studies to investigate the effect of components of the proposed BiFSMN, including the High-frequency Enhancement Distillation (HED) and Thinnable Binarization Architecture (TBA), on the Speech Commands V1-12 and V2-12 KWS tasks. 

As shown in Table ~\ref{tab:ablation}, the vanilla binarization baseline suffers a significant performance drop in both datasets. The naive distillation scheme helps with the accuracy on the basic D-FSMN architecture, and the application of HED further improves the performance considerably based on distillation. It demonstrates that emphasizing high-frequency information makes it easier for the binarized network to exploit the most crucial representation. And the distillation strategy is also indispensable for better aligning and transferring information. 

On the other hand, when solely utilizing TBA, we train a single model but optimize it in different parameter scales through weighted backward propagation. It shows a great deal of potential in not only the adaptive and lightweight computation at runtime but also the optimization of the binarized network during training. 
Moreover, jointly using HED and TBA further close the accuracy gap between the binarized model and the full-precision counterpart, which is less than 3\% on both datasets.

\begin{table}[t]
\tablesize
\centering
% \vspace{-0.55in}
\renewcommand\arraystretch{0.4}
% \caption{Comparison of SOTA binarization methods on Speech Commands V1 and V2 datasets.}
% \vspace{-0.15in}
% \setlength{\tabcolsep}{2.0mm}
{\begin{tabular}{llcrrr}
\toprule
{\bf Dataset} & {\bf Method}  & {\tabincell{r}{\textbf{\#Bits}}} & {\tabincell{r}{\textbf{FLOPs}}}  & {\tabincell{r}{\textbf{V1}}} & {\tabincell{r}{\textbf{V2}}}\\
\midrule
\multirow{15}{*}{\tabincell{l}{Speech\\Commands\\12}} & Full Prec. & 32/32 & 710.15 & 97.93 & 98.05\\
\cmidrule{2-6}
& DoReFa  & 1/1 & 40.46 & 66.42 & 66.59\\
& BNN     & 1/1 & 35.87 & 68.84 & 70.87\\
& RAD     & 1/1 & 35.87 & 71.51 & 69.80\\
& XNOR    & 1/1 & 45.04 & 82.74 & 87.34\\
& Bi-Real & 1/1 & 40.46 & 85.87 & 87.93\\
& IR-Net  & 1/1 & 40.46 & 86.81 & 85.10\\
\cmidrule{2-6}
& \multirow{3}{*}{\tabincell{l}{BiFSMN\\$_{[1, 0.5, 0.25]\times}$}}    & \multirow{3}{*}{1/1} & 40.46 & \textbf{95.03} & \textbf{94.86} \\
& & & 29.90 & 94.87 & 94.73\\
& & & \textbf{24.62} & 94.48 & 94.63\\
\midrule
\multirow{10}{*}{\tabincell{l}{Speech\\Commands\\20}} & Full Prec.    & 32/32 & 711.20 & 96.57 & 97.00\\
\cmidrule{2-6}
& XNOR    & 1/1 & 45.04 & 80.69 & 85.05\\
& Bi-Real & 1/1 & 41.50 & 80.84 & 84.39\\
& IR-Net  & 1/1 & 41.50 & 83.78 & 83.32\\
\cmidrule{2-6}
& \multirow{3}{*}{\tabincell{l}{BiFSMN\\$_{[1, 0.5, 0.25]\times}$}}    & \multirow{3}{*}{1/1} & 41.50 & \textbf{92.88} & \textbf{92.98} \\
& & & 30.95 & 92.67 & 92.81\\
& & & \textbf{25.67} & 92.65 & 92.72\\
\midrule
\multirow{10}{*}{\tabincell{l}{Speech\\Commands\\35}} & Full Prec.    & 32/32 & 713.16 & 96.63 & 95.96 \\
\cmidrule{2-6}
& IR-Net  & 1/1 & 43.47 & 74.09 & 74.93\\
& Bi-Real & 1/1 & 43.47 & 80.86 & 81.86\\
& XNOR    & 1/1 & 48.06 & 81.25 & 84.05\\
\cmidrule{2-6}
& \multirow{3}{*}{\tabincell{l}{BiFSMN\\$_{[1, 0.5, 0.25]\times}$}} & \multirow{3}{*}{1/1} & 43.47 & \textbf{92.10} & \textbf{90.67}\\
& & & 32.91 & 91.93 & 90.54\\
& & & \textbf{27.63} & 91.85 & 90.42\\
\bottomrule
\end{tabular}}
\caption{Comparison of SOTA binarization methods on Speech Commands V1 and V2 datasets.}
\label{tab:methods}
\end{table}

\subsection{Comparative Experiments}

% We first compare our BiFSMN with existing binarization methods which are applicable to various structures, including BNN, DoReFa, XNOR, Bi-Real, IR-Net, RAD. And the existing binarization methods are evaluated on 8-block binarized D-FSMN architecture with the same size as the largest option of BiFSMN. 
We first compare our BiFSMN with existing structure-independent binarization methods, including BNN~\cite{BNN}, DoReFa~\cite{dorefa}, XNOR~\cite{XNOR}, Bi-Real~\cite{BiReal}, IR-Net~\cite{qin2019forward}, and RAD~\cite{Regularize-act-distribution}. We evaluate these binarization methods on 8-block D-FSMN architecture with the same size as the largest variant of BiFSMN. 
The results in Table~\ref{tab:methods} show that our 1-bit BiFSMN completely outperforms other SOTA binarization methods by a wide margin. %on the same architecture. On Speech Commands V1-12 tasks, BiFSMN surpasses the XNOR and IR-Net even 15.19\% and 20.57\%, respectively. 
% And BiFSMN even enjoys the competitive accuracy compared to full-precision counterparts on several tasks, e.g., it only drops x\% accuracy in Speech Command V1-12 task. 
% Furthermore, the thinner version BiFSMN$_{.5\times}$ with 4 blocks and BiFSMN$_{.25\times}$ with 2 blocks even achieves xx$\times$ and xx$\times$ saving in GFLOPs yet impressive xx\% and xx\% accuracy, respectively. 
It is noteworthy that BiFSMN even enjoys competitive accuracy to full-precision counterparts within 4\% average accuracy drop on both datasets. %\textit{e.g.}, it only drops 2.9\% accuracy in Speech Command V1-12 task. 
%Furthermore, the thinner version BiFSMN$_{0.5\times}$ with 4 blocks and BiFSMN$_{0.25\times}$ with 2 blocks even achieves 23.8$\times$ and 28.8$\times$ FLOPs saving without sacrificing accuracy (only 0.16\% and 0.13\% drop) on Speech Commands V1-12 task. 

% Moreover, to validate the advantage of our method from the architecture perspective, we also compare the binarization on various networks in Table~\ref{tab:architectures}, including FSMN, VGG, BC-ResNet, and Audiomer that are widely used in KWS.
% As our base architecture, the binarized D-FSMN far outperforms other binarized architectures, indicating that it is binarization-friendly. For example, when applying the same XNOR method, binarized D-FSMN higher xx\% than binarized VGG16bn with more parameters and FLOPs.
Second, to validate the advantage of our TBA from the architecture perspective, we also compare it with various networks widely used in KWS, including FSMN~\cite{zhang2015feedforward}, VGG19bn~\cite{VeryDeepConvolutional}, BC-ResNet~\cite{kim2021broadcasted}, and Audiomer~\cite{sahu2021audiomer}. We binarized these architectures with XNOR and IR-Net. In Table~\ref{tab:architectures}, our HED can generally be applied in FSMN-based architectures and make a difference for the binarized model performance. 
Moreover, equipped with TBA, BiFSMN can further strike a balance between accuracy and efficiency at runtime. 
% (editor) The thinner version BiFSMN$_{0.5\times}$ with 4 blocks and BiFSMN$_{0.25\times}$ with 2 blocks even achieve 23.8$\times$ and 28.8$\times$ FLOPs saving without sacrificing accuracy (only 0.16\% and 0.13\% drop) on Speech Commands V1-12 task. 
%The BiFSMN$_{0.5\times}$ and BiFSMN$_{0.25\times}$ variants reduce the computation FLOPs by 17.6$\times$ and 28.8$\times$ respectively. %but maintain excellent performance (with only 0.16\% and 0.55\% accuray drop). %reduces the parameter amount and computation FLOPs in inference
We further prune the model width and provide an extremely tiny BiFSMN$_\textrm{S}$ (with 32 backbone memory size and 64 hidden size) with only 0.05M parameters and 9.16M FLOPs, %with trivial cost in model accuracy (about 5\%), 
demonstrating that our methods also works well on tiny networks.

\begin{table}[t]
\tablesize
\centering
% \vspace{-0.4in}
\renewcommand\arraystretch{0.4}
% \caption{Comparison of various binarized architectures for KWS on Speech Commands V1-12 task.}
% \vspace{-0.15in}
% \setlength{\tabcolsep}{1.2mm}
{\begin{tabular}{llcrrr}
\toprule
{\bf Arch.} & {\bf Quant}  & {\tabincell{c}{\textbf{\#Bits}}} & {\tabincell{r}{\textbf{\#Param}}}  & {\tabincell{r}{\textbf{FLOPs}}}  &{\bf \tabincell{r}{Acc.}} \\
\midrule
\multirow{4}{*}{\tabincell{l}{VGG19bn}} & Full Prec.    & 32/32 & 38.97 & 53348.40 & 97.74\\
\cmidrule{2-6}
& IR-Net & 1/1 & 38.97 & 1030.18 & 63.86\\
& XNOR & 1/1 & 38.97 & 1061.63 & 66.10\\
\midrule
\multirow{4}{*}{\tabincell{l}{BC-ResNet}} & Full Prec. & 32/32 & 0.35 & 3749.71 & 97.84\\
\cmidrule{2-6}
& XNOR & 1/1 & 0.35 & 619.03 & 65.94\\
& IR-Net & 1/1 & 0.35 & 541.70 & 66.51\\
\midrule
\multirow{4}{*}{\tabincell{l}{Audiomer}} & Full Prec. & 32/32 & 0.80 & 18577.49 & 98.95\\
\cmidrule{2-6}
& IR-Net & 1/1 & 0.80 & 1449.41 & 62.44\\    
& XNOR & 1/1 & 0.80 & 2567.19 & 62.44\\
\midrule
\multirow{6}{*}{\tabincell{l}{FSMN}} & Full Prec. & 32/32 & 0.45 & 1625.29 & 97.52\\
\cmidrule{2-6}
& XNOR & 1/1 & 0.45 & 80.61 & 55.89\\
& IR-Net & 1/1 & 0.45 & 64.36 & 88.35\\
& HED (ours) & 1/1 & 0.45 & 64.36 & \textbf{89.90}  \\
\midrule
\multirow{6}{*}{\tabincell{l}{D-FSMN}} & Full Prec. & 32/32 & 0.60 & 710.15 & 97.93\\
\cmidrule{2-6}
& XNOR & 1/1 & 0.60 & 45.04 & 82.74\\
& IR-Net & 1/1 & 0.60 & 40.46 & 86.81 \\
& HED (ours) & 1/1 & 0.60 & 40.46 & \textbf{93.47} \\
\midrule
\multirow{9}{*}{\tabincell{l}{BiFSMN\\$_{[1, 0.5, 0.25]\times}$}}  & Full Prec. & 32/32 & 0.60 & 710.15 & 97.93 \\
\cmidrule{2-6}
& \multirow{3}{*}{\tabincell{l}{IR-Net}} & \multirow{3}{*}{1/1} & \multirow{3}{*}{0.61} & 40.46 & 88.37 \\
& & & & 29.90 & 87.11 \\
& & & & 24.62 & 86.08 \\
\cdashline{2-6}[1pt/1pt]
 & \multirow{3}{*}{HED (ours)} & \multirow{3}{*}{1/1} & \multirow{3}{*}{0.61} & 40.46 & \textbf{95.03}\\
 & & & & 29.90 & 94.87\\
 & & & & \textbf{24.62} & 94.48\\
\midrule
\multirow{9}{*}{\tabincell{l}{BiFSMN\\$_{[1, 0.5, 0.25]\times}$}}  & Full Prec. & 32/32 & 0.05 & 91.62 & 97.51 \\
\cmidrule{2-6}
& \multirow{3}{*}{\tabincell{l}{IR-Net}} & \multirow{3}{*}{1/1} & \multirow{3}{*}{0.05} & 11.94 & 72.21 \\ 
& & & & 10.08 & 71.70 \\ 
& & & & 9.16 & 71.66 \\ 
\cdashline{2-6}[1pt/1pt]
& \multirow{3}{*}{HED (ours)} & \multirow{3}{*}{1/1} & \multirow{3}{*}{0.05} & 11.94 & \textbf{90.65}\\
 & & & & 10.08 & 90.33\\
 & & & & \textbf{9.16} & 90.31\\
\bottomrule
\end{tabular}}
\caption{Comparison of various binarized architectures for KWS on Speech Commands V1-12 task. \#Param denotes million parameters. }
\label{tab:architectures}
\end{table}

\subsection{Deployment Efficiency}

% (editor) As mentioned, low memory footprint and fast real-time response are highly expected in KWS while running on real-world edge devices. 
To validate the practicability of BiFSMN, we test the actual speed of BiFSMN on Raspberry Pi 3B+ with 1.2GHz 64-bit ARMv8 CPU Cortex-A53.

\begin{figure}
\centering
% \vspace{-0.18in}
\includegraphics[width=8.cm]{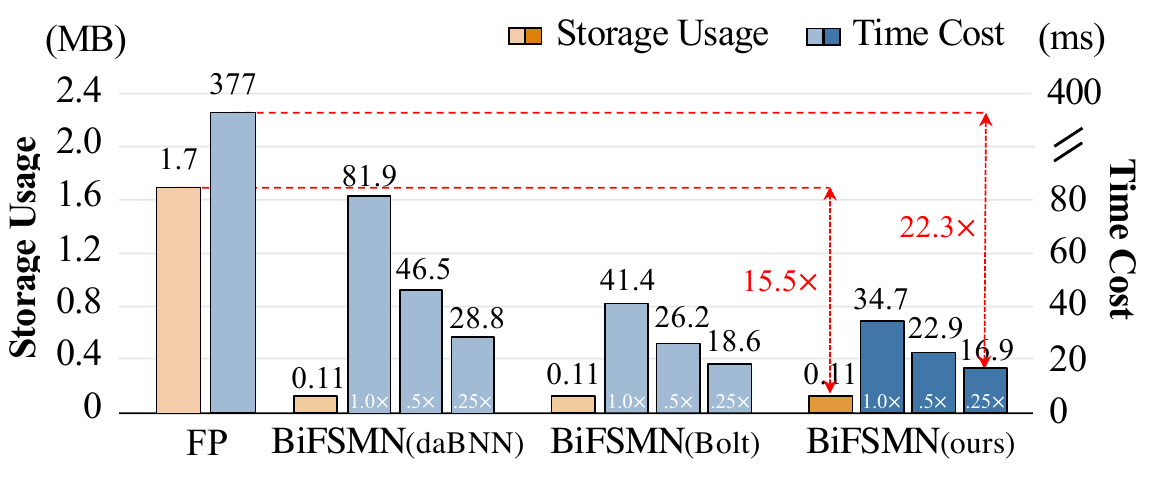}
% \vspace{-0.18in}
\caption{Performance evaluation on real-world ARMv8 devices.}
\label{fig:arm_performance}
\end{figure}

According to Figure~\ref{fig:arm_performance}, due to the proposed optimized 1-bit Fast Bitwise Computation Kernel, our BiFSMN delivers $10.9\times$ acceleration compared to the full-precision counterparts. It is also much faster than the existing open-source high-performance binarization frameworks (daBNN and Bolt). 
Furthermore, benefiting from the thinnable architecture, BiFSMN can adaptively balance accuracy and efficiency at runtime according to the resources on device, and switches to BiFSMN$_{0.5\times}$ or BiFSMN$_{0.25\times}$ for further $15.5\times$ and $22.3\times$ speedups, respectively. It shows that our BiFSMN can satisfy different resource constraints. % size and computation cost can be extremely compressed. It is noteworthy that the thinned 1-bit BiFSMN$_{0.25\times}$ can achieve $22.3\times$ speedup compared with the full-performance network. 

\section{Conclusion}
We present BiFSMN, an accurate and extreme-efficient binary neural network for KWS. We first construct an HED scheme to emphasize high-frequency information to optimize the training of the binarized network. We also propose a TBA to achieve instant and adaptive accuracy-efficiency trade-offs at runtime. BiFSMN outperforms existing binarization methods by convincing margins and is even comparable to the full-precision counterpart. Moreover, our implementation for BiFSMN on ARMv8 real-world devices achieves an impressive $22.3\times$ speedup and $15.5\times$ storage-saving. 
% (editor) Our work demonstrates the great potential of binarization for KWS, and we hope our work will inspire future research.

% \
% \noindent\textbf{Acknowledgement} This work was supported in part by National Natural Science Foundation of China under Grant 62022009 and Grant 61872021, Beijing Nova Program of Science and Technology under Grant Z191100001119050.

\section*{Acknowledgements}
\fontsize{12}{0}This work was supported in part by National Natural Science Foundation of China under Grant 62022009 and Grant 61872021, Beijing Nova Program of Science and Technology under Grant Z191100001119050.

\clearpage
{
\bibliographystyle{named}
\bibliography{ijcai22}
}

\end{document}